\documentclass{bmvc2k}

\usepackage{times}
\usepackage{epsfig}
\usepackage{graphicx}
\usepackage{amsmath}
\usepackage{amssymb}
\usepackage{amsfonts}
\usepackage{bm}
\usepackage{float}
\usepackage[inline]{enumitem}
\usepackage{booktabs}
\usepackage{diagbox}
\usepackage{multirow}
\usepackage{multicol}
\usepackage{mwe}
\usepackage{soul}
\usepackage{lipsum}
\usepackage{pifont}
\usepackage{bbm}
\usepackage{booktabs}
\usepackage[ruled,vlined,commentsnumbered,linesnumbered]{algorithm2e}

\def\eqref#1{equation~\ref{#1}}

\def\1{\bm{1}}

\DeclareMathAlphabet{\mathsfit}{\encodingdefault}{\sfdefault}{m}{sl}
\SetMathAlphabet{\mathsfit}{bold}{\encodingdefault}{\sfdefault}{bx}{n}

\title{Two is a crowd: tracking relations in videos}

\addauthor{Artem Moskalev}{a.moskalev@uva.nl}{1}
\addauthor{Ivan Sosnovik}{i.sosnovik@uva.nl}{1}
\addauthor{Arnold Smeulders}{a.w.m.smeulders@uva.nl}{1}

\addinstitution{
 UvA - Bosch Delta Lab\\
 University of Amsterdam, Netherlands\\
 Amsterdam, Netherlands
}

\runninghead{Moskalev, Sosnovik, AND Smeulders}{Tracking relations in videos}

\def\etal{\emph{et al}\bmvaOneDot}

\begin{document}

\maketitle

\begin{abstract}

Tracking multiple objects individually differs from tracking groups of related objects. When an object is a part of the group, its trajectory depends on the trajectories of the other group members. Most of the current state-of-the-art trackers follow the approach of tracking each object independently, with the mechanism to handle the overlapping trajectories where necessary. Such an approach does not take inter-object relations into account, which may cause unreliable tracking for the members of the groups, especially in crowded scenarios, where individual cues become unreliable due to occlusions. To overcome these limitations and to extend such trackers to crowded scenes, we propose a plug-in Relation Encoding Module (REM). REM encodes relations between tracked objects by running a message passing over a corresponding spatio-temporal graph, computing relation embeddings for the tracked objects. Our experiments on MOT17 and MOT20 demonstrate that the baseline tracker improves its results after a simple extension with REM. The proposed module allows for tracking severely or even fully occluded objects by utilizing relational cues.
\end{abstract}

\begin{figure*}[ht!]
  \centering
    \includegraphics[width=0.97\linewidth]{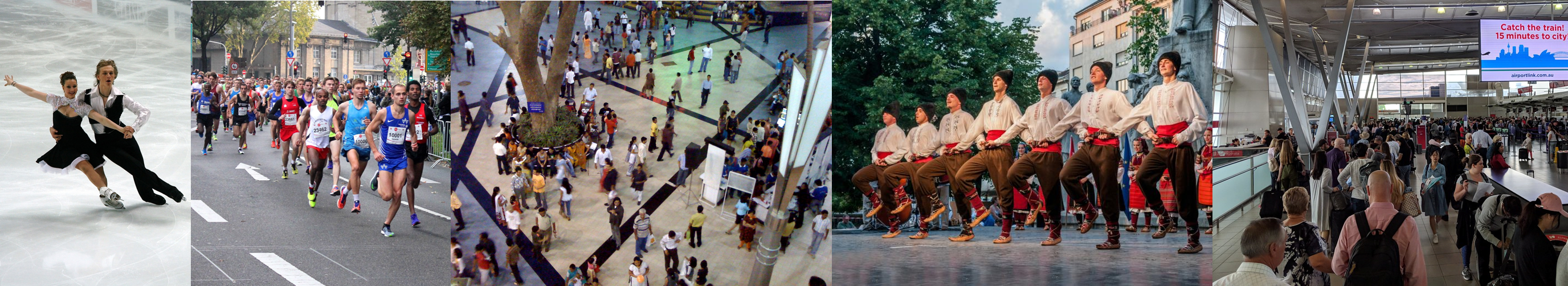}
    \caption{In dancing, shopping malls, or crowds, dense overlap between individuals makes tracking hard. In this paper we keep track of relations among individuals as a goal by itself, and to better track each target as a member of a group. Effectively, we aim to employ heavy and persistent overlap with the partner rather than rejecting that information.}
  \label{fig:toc}
\end{figure*}


\vspace{-6mm}
\section{Introduction}
\label{sec:intro}

For online multi-object tracking, when objects are part of a group, the frequent mutual occlusions make individual tracking harder. Rather than rejecting that information, identifying group membership is interesting by itself, where in principle the group is easier to identify having more uniquely identifying characteristics than an individual object would. In this paper we set out to exploit group relations for multi-object tracking.

When tracking pedestrians online in a crowd, following one specifically is generally harder than following all members of a family of three, just because their \textit{combination} offers good distinction: one tall with one small person, each with a trolley. Occlusion may hamper complete view of one of the targets but then the characteristics of related members may be borrowed to render approximate tracking for the occluded one like parents with a child in shopping malls and other forms of crowd control. See Figure \ref{fig:toc} for other examples of dense interaction, where tracking group relations is advantageous. 

\begin{figure*}[t!]
  \centering
    \includegraphics[width=0.95\linewidth, height=4cm]{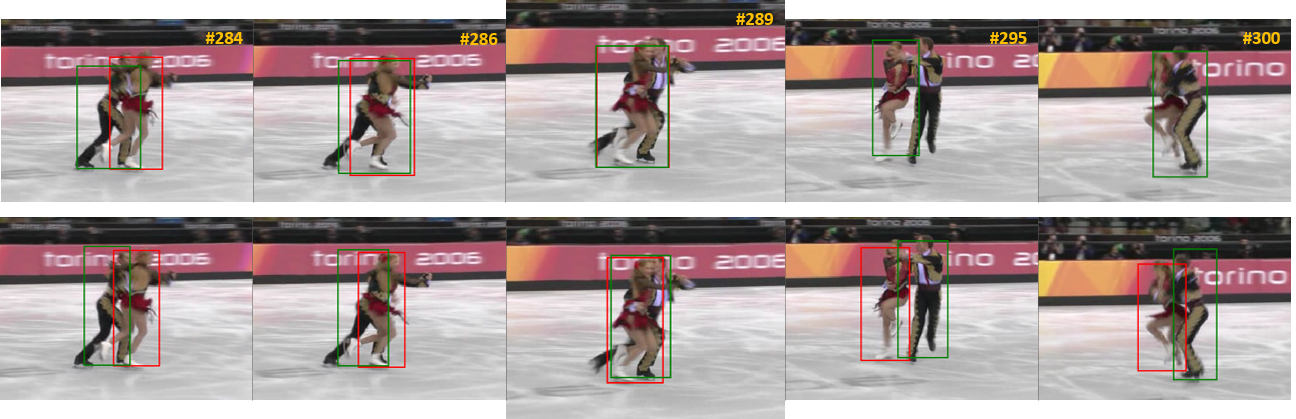}
  \caption{Top: multi-object tracking with Tracktor \cite{tracktor_2019_ICCV}, where independent trajectories are assumed making tracking hard when the targets come close. Bottom: the same \cite{tracktor_2019_ICCV} tracker equipped with relation-reasoning is more robust to occlusions.}
  \vspace{-3mm}
  \label{fig:fig1intro}
\end{figure*}

Multi-object online tracking has recently made great progress with tracking-by-regression \cite{tracktor_2019_ICCV, zhou2020tracking, zhang2020fair, Lu_2020_CVPR, wang2019towards, xu2020train, Voigtlaender19CVPR_MOTS}. These methods track each object separately until, at a crossroad of tracks, a mechanism is called upon to determine which object continues on what track. The current methods demonstrate good speed and good accuracy. They do not, however, consider inter-object relations, which may cause tracking to become unreliable especially when the interaction between bodies becomes dense where occlusion becomes a major obstacle, as in (Figure \ref{fig:fig1intro}).

We draw inspiration from multi-object processing, where the whole video is available for the analysis. In \cite{tang2015subgraph, tang2016multi, keuper2016multicut, keuper2020jcc, tang2017multiple}, the trajectories are derived by running a graph optimization on the object detections. While the structure of the graph encodes the inter-object relations in these offline trackers, their capability of finding relations relies heavily on having all detections in the video at once, combining information \textit{before and after} dense interactions. This offline information blocks the methods unsuited for online multi-object tracking as the detections of the future are not yet available.

In this work, we extend current tracking-by-regression methods with online group relations. Inspired by offline graph-based video analysis, we learn to encode inter-object relations from limited data \textit{a priory}. In our relation encoding module, a message passing algorithm is running over a dynamic object-graph to produce relation embeddings, which encode the group structure for each object. The message passing is performed over space and time to capture the complex dynamics of group behaviour. The relation encoding module is implemented as a plug-in extension for tracking-by-regression methods.

We make the following contributions:
\begin{itemize}
    \item We develop a method to encode inter-object relations online in dense scenes by running spatial-temporal message passing. 
    \item We demonstrate the virtue of relations to improve the tracking of objects in dense scenes by adding relations on top of current tracking-by-regression methods, even tracking objects with low visibility.
    \item And we demonstrate an online view of the degree of the relationships between objects.
\end{itemize}
\section{Related work}
\label{sec:related}

\paragraph{Multi-object tracking by graph association}

Many the multi-object trackers first apply an object detector on the whole sequence, then link the detections across frames on the basis of a best match criterion \cite{tang2015subgraph, tang2016multi, keuper2016multicut, keuper2020jcc, tang2017multiple}. They follow the tracking by a graph association paradigm. The matching is usually posed as an offline graph association problem connecting the detections into trajectories. In \cite{tang2015subgraph, tang2016multi}, the authors solve association as a multicut problem, where trajectories are derived from a dense graph of detections by extracting weighted subgraphs. Along the same lines, in \cite{keuper2016multicut}, Keuper \etal propose a multicut formulation to decompose a dense detection graph into a set of trajectories. To better handle occlusions, Tang \etal \cite{tang2017multiple} further extend multicuts with lifted edges.

Graph associations are powerful as they reason about groups of detections, while taking inter-object relations into account. However, their offline nature limits the real-life application of such methods. Also, due to their combinatorial non-differentiable formulation, it is not trivial to combine these graph association algorithms with modern end-to-end trackers. In this work, we take inspiration from these offline graph association works and develop a new method for relation encoding, which learns to encode the dynamics of multiple objects for online tracking. Our relation encoding module is fully end-to-end compatible with modern trackers.

\paragraph{Multi-object tracking by regression association}

Recently, a family of methods called tracking-by-regression has become the state-of-the-art approach in multi-object tracking. The key idea is to assess the association of detections to previously detected objects by utilizing the regression head of the object detector. In the pioneering work of Bergmann \etal \cite{tracktor_2019_ICCV}, tracking is based on the second stage of the Faster R-CNN \cite{ren2017faster} object detector with the previous positions of detected objects as proposals. Later, more sophisticated object detectors were used \cite{zhou2020tracking, Lu_2020_CVPR, zhang2020fair, wang2019towards}. In \cite{zhou2020tracking} Zhou \etal modify CenterNet \cite{zhou2019objects} for multi-object tracking. In \cite{Lu_2020_CVPR}, authors modify the lightweight RetinaNet \cite{Detectron2018} for faster inference. In \cite{zhang2020fair, wang2019towards}, the authors extend the detector with ReID embeddings, which allows for better identity preservation in case of occlusion. 

In all these works, objects are tracked independently of one another. When scenes become crowded or filled with similar targets, independent tracking under frequent and heavy occlusion becomes hard or impossible. Whereas the above methods function well generally, they tend to break when individual cues are no longer available (Figure \ref{fig:fig1intro}). For these hard but frequent circumstances, one has to employ relations in tracking. In this paper, we propose a simple yet effective extension for regression-based multi-object trackers to improve tracking robustness in dense interaction.

\paragraph{Reasoning on relations}

From the literature on relations in computer vision, we focus in particular on self-attention \cite{Vaswani2017attention, lee2019set} and graph neural networks (GNN) \cite{wu2021gnnsurvey, velickovic2018graph}. Their ability to use structural information in the input has motivated Hu \etal \cite{Hu2018reldet} to develop a self-attention relation module to remove duplicates in the task of object detection. In \cite{cai2019exploiting} Cai \etal use graph convolution to propagate relational information to refine a predicted pose. Narasimhan \etal \cite{Narasimhan2018OutOT} rely on graph-structured representations to encode relations for visual question answering. In \cite{materzynska2020somethingelse}, Materzynska \etal model interactions in a subject-object graph representation for action recognition.

These works demonstrate how relation cues improve the analysis. Inspired by the success of relation reasoning for various tasks, we expand relation cues to multi-object online tracking as a plug-in extension for state-of-the-art trackers and with the potential online applicability for all above purposes.

\paragraph{Metrics of multi-object tracking} 

Classical multi-object tracking evaluation methods include CLEAR MOT metrics \cite{bernardin2008clearmot} and IDF1 score \cite{ristani2016idf}. These metrics assess various aspects of tracking, while the recently proposed HOTA metric \cite{luiten2020IJCV} aims to provide the measure of overall performance. None of these metrics, however, capture, the dense interaction of objects, which is one of the circumstances which makes multi-object tracking interesting. To do so, in this work we also consider the decomposition of the HOTA metric over various localization thresholds.
\section{Encoding relations}
\label{sec:method}

\begin{figure*}[t]
  \centering
    \includegraphics[width=0.95\linewidth]{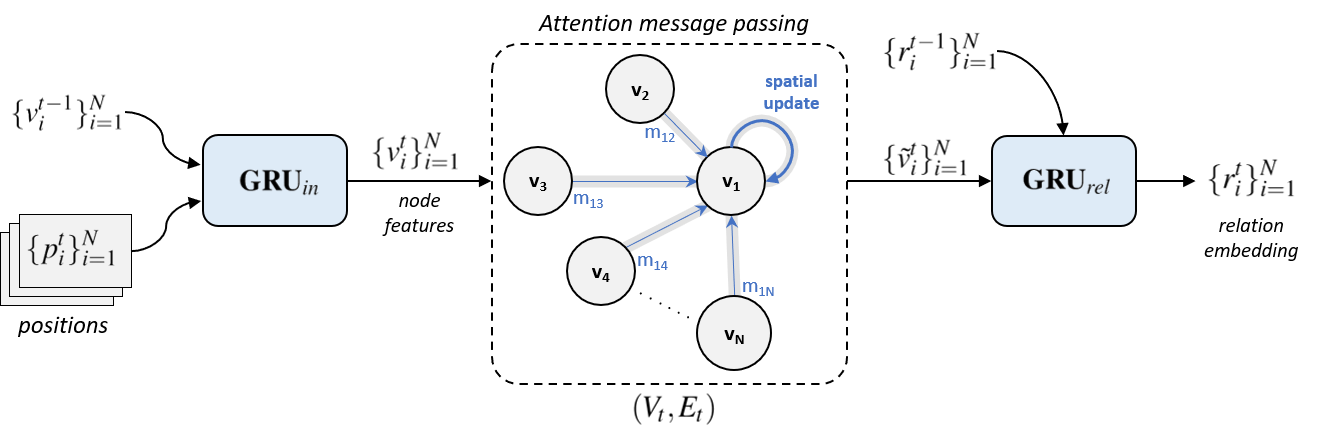}
  \caption{Computing relation embeddings $\{r_{i}^{t} \}_{i=1}^N$ for N object from time step $t-1$ to $t$. Input coordinates are passed through an input GRU-cell to produce node features. Message passing is performed on top of the relational graph to update internal node representations. Finally, node representations are passed through another GRU-cell, which emulates message passing along temporal edges.}
  \label{fig:fig2rem}
\end{figure*}

To encode inter-object relations, the relation encoding module takes a set of tracked instances as input and produces \textit{relation embeddings} by running a message passing algorithm over the spatial-temporal graph. Figure \ref{fig:fig2rem} renders the architecture of the module.

\subsection{Building relational graph}

We define $G_T = \{ (V_t, E_t)_{t=1}^T ; (Z_t)_{t=1}^{T-1} \}$ as a spatial-temporal graph, where $T$ is a total number of time steps, $V_t, E_t$ represent the vertices and edges of the graph at the time step $t$, respectively. $Z_t$ is a set of temporal edges from $t$ to $t+1$. Vertices correspond to the objects as tracked, while the temporal edges encode their trajectories. Only the nodes, which correspond to the same instance, are linked in time. To decide on the spatial edges at time step $t$, we first compute the distance matrix $D^{t}$ with entries:

\begin{equation}
    D_{ij}^{t} = \sqrt{\frac{(x_{i}^{t} - x_{j}^{t})^2}{\Bar{w}_{ij}^{t}} - \frac{(y_{i}^{t} - y_{j}^{t})^2}{\Bar{h}_{ij}^{t}}}
\end{equation}

where $(x_{i}^{t}, y_{i}^{t}, w_{i}^{t}. h_{i}^{t})$ corresponds to the center coordinates, width and height of the $i$-th object and $\Bar{w}_{ij}^{t}=\min (w_{i}^{t}, w_{j}^{t})$ respectively. We use the scaled Euclidean distance to prevent linking remote instances, which may be close if evaluated only by the center coordinates, but far away in depth. To obtain an adjacency matrix $A$ we simply threshold the distances, i.e. $A_{ij}^{t} = \mathbbm{1} [D_{ij}^{t} \leq d_{th}]$, where $d_{th}$ is a hyper-parameter.

\subsection{Graph-attention message passing}

Inter-object relations are modulated by running message passing over the relational graph. The procedure consists of 4 steps: compute input node features, compute messages between spatial nodes, aggregate messages and compute spatial-temporal updates of node representations. This procedure is recurrently performed for each time step until the end of the graph is reached.

\paragraph{Node features}

To construct the input feature we use bounding box coordinates of the detection and the positional offset with respect to the previous time step. Let $p_{i}^{t} \in \mathbbm{R}^{4}$ be the bounding box of the \textit{i-th} object at time $t$, the input feature $v_{i}^{t} \in \mathbbm{R}^{F}$ for the node is then computed as:
\vspace{-1mm}
\begin{gather}
    \Tilde{p}_{i}^{t} =  \sigma(\textbf{W}_{in} [p_{i}^{t} \| p_{i}^{t} - p_{i}^{t-1}] + \textbf{b}_{in}) \\
    v_{i}^{t} = \textbf{GRU}_{in}(\Tilde{p}_{i}^{t}, v_{i}^{t-1})
\end{gather}

where $\textbf{W}_{in}, \textbf{b}_{in}$ are learnable parameters, $\sigma$ is a non-linearity and $\|$ denotes concatenation operator. The initial hidden states of the $\textbf{GRU}_{in}$ cell are set to zeros.

\paragraph{Message sending}

A message between two nodes of the graph is designed to encode their pairwise interaction. We define the message as a function of both the sending and the receiving nodes $i$ and $j$, respectively. To make the message aware of the geometry of the graph, we also include the distance $D_{ij}^{t}$ between the objects as an additional input for the message function. The message $m_{ij}^{t}:\mathbbm{R}^{F} \times \mathbbm{R}^{F} \times \mathbbm{R} \rightarrow \mathbbm{R}^{F}$ is calculated as: 

\begin{equation}
    m_{ij}^{t} = \sigma \big{(}\textbf{W}_{m_2} (\sigma (\textbf{W}_{m_1} [ v_{i}^{t} \| v_{j}^{t} \| D_{ij}^{t}] + \textbf{b}_{m_1})) + \textbf{b}_{m_2} \big{)}
\end{equation}

\paragraph{Aggregating messages} 

When the messages have been computed, they are gathered in an aggregated message. An aggregation function should be permutation equivariant with respect to the neighbors' features. In this work, we follow the graph attention approach \cite{velickovic2018graph}, which computes attention between features to weigh them according to their importance. The attention mechanism $\alpha_{ij}^{t}: \mathbbm{R}^{F} \times \mathbbm{R}^{F} \rightarrow \mathbbm{R}_{+}$ computes the attention coefficients as:

\begin{equation}
\label{eq:alpha_att}
    \alpha_{ij}^{t} = \frac{\exp(\text{LeakyReLU} \big{(} [\textbf{W}_{a_1} v_{i}^{t}]^{T} [\textbf{W}_{a_2} v_{j}^{t}] \big{)})}{\sum_{j \in \mathcal{N}_i} \exp(\text{LeakyReLU} \big{(} [\textbf{W}_{a_1} v_{i}^{t}]^{T} [\textbf{W}_{a_2} v_{j}^{t}] \big{)})}
\end{equation}

where $\mathcal{N}_i$ denotes the set of the nodes \textit{spatially} adjacent to \textit{i-th} node in the graph. Temporal edges are not considered at this stage. The  attention coefficients are then used to compute a linear combination of the corresponding neighbors' representation into an aggregated feature.

\paragraph{Spatial-Temporal update}

In the final step, we update node representations spatially and temporally. For the spatial update, we concatenate the self-feature of the node with the aggregated message from its neighbors and pass it through the perceptron. The temporal update is performed by passing the features through the GRU-cell. Formally:

\begin{align}
\label{eq:st_update}
     \hspace{13mm} \textit{(spatial)} & \hspace{5mm} \Tilde{v}_{i}^{t} = \sigma(\textbf{W}_{u}[v_{i}^{t} \| \sum_{j \in \mathcal{N}_i} \alpha_{ij}^{t} m_{ij}^{t}] + \textbf{b}_{u}) && \\
     \label{eq:st_update2}
    \hspace{13mm} \textit{(temporal)} & \hspace{5mm} r^{t}_i = \textbf{GRU}_{rel} (\Tilde{v}_{i}^{t}, r_{i}^{t-1}) &&
\end{align}

We call the resulting feature $r^{t}_i \in \mathbbm{R}^{F}$ \textit{relation embedding} of the \textit{i-th} node at time \textit{t}. Relation embeddings at $t=0$ are all set to zero vectors.

As can be seen in Equations \ref{eq:st_update}, \ref{eq:st_update2}, in our implementation the temporal updates follow the spatial update. Early experiments demonstrated that such an approach is slightly superior to when the temporal update is performed first (data not shown).

\subsection{Relation-importance}
To answer the question \textit{to what degree object $i$ relates to object $j$ at time $t$}, we define a relation-importance function $R^{t}_{ij} : \mathbbm{R}^{F} \times \mathbbm{R}^{F} \rightarrow \mathbbm{R}_{+}$:

\begin{equation}
    \label{eq:rel_imp}
    R^{t}_{ij} = \mathbbm{1} [D_{ij}^{t} \leq d_{th}] \phi (r_{i}^{t}, r_{i \oslash j}^{t})
\end{equation}

where $\phi$ can be any bounded metric and $r_{i \oslash j}^{t}$ denotes the relation encoding of \textit{i-th} object computed by excluding the \textit{j-th} node from the set of neighbors. In early experiments, we found $\phi(x, y) = 1 - cos^2(x, y)$ to work well as a metric (data not shown).

Higher values for $R^{t}_{ij}$ indicate a higher degree of relation between these objects. Note that we can also compute relation-importance by adopting the attention weights from Equation \ref{eq:alpha_att}. However, we observed that such an approach yields less intuitive relations as it does not take the information encoded in the messages into account. In addition, it restricts the relations to be \textit{symmetric}, while formulation from Equation \ref{eq:rel_imp} permits \textit{asymmetric} relations.


\subsection{Utilizing relations for tracking}
\label{sec:model}

Next, we explore two purposes of using relations in tracking: \textit{(i)} extending tracking-by-regression models to reason about the object's position based both on appearance and relation cues, \textit{(ii)} recovering the position of the objects purely from their relation embeddings, which is useful in the case of occlusions.

\paragraph{Relation-aware tracking-by-regression}

To make a tracker aware of relations, we condition the predicted positions of the objects on their relation embeddings. To that end, we concatenate the appearance features extracted from proposal regions with the relation embeddings of the corresponding objects. The positional offset is then predicted by passing the combined feature via the regression head of the object detector. The model can be seen as the REM with the tracker attached to graph nodes. Such a framework applies to a wide range of trackers \cite{tracktor_2019_ICCV, zhou2020tracking, zhang2020fair, Lu_2020_CVPR, wang2019towards}. It does not require modification of the tracker other than adjusting the regression head, see the supplementary material.

\paragraph{Tracking-by-relations only}
When the object is heavily occluded, appearance features become unreliable, causing a tracking failure. As relation embeddings produced by the relation module do not depend on the objects' appearance, they can be used to track objects under severe occlusion. To that end, we train an output MLP to directly regress the coordinates of the occluded objects from their relation embeddings, see the supplementary material. Then, we run the relation-aware tracker on the whole sequence and apply tracking-by-relations only to the occluded objects, when the rest of the objects are tracked both based on appearance and relational cues. This way, the model can \textit{explain} hard occluded cases through easy non-occluded ones. 
\begin{figure*}[t]
\vspace{1mm}
  \centering
    \includegraphics[width=0.98\linewidth]{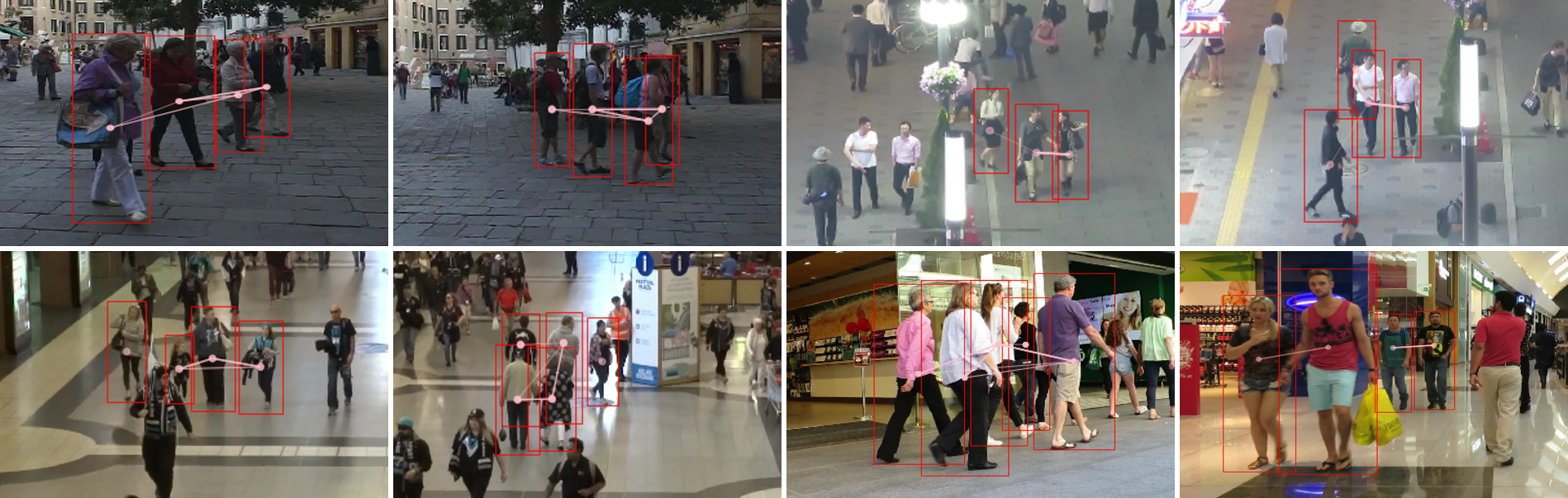}
  \caption{A visualization of relation-importance weights during the tracking on the MOT17/MOT20 benchmarks.}
  \label{fig:discovered_rel}
\end{figure*}

\section{Experiments}
\label{sec:experiments}

We evaluate relation-aware tracker on the MOTChallenge benchmarks MOT17 \cite{Milan2016MOT16AB} and MOT20 \cite{dendorfer2020mot20}. 

\subsection{Implementation details}

We use Tracktor\footnote{\url{https://github.com/phil-bergmann/tracking_wo_bnw}} as our baseline model as it provides good speed-accuracy balance. We extend Tracktor to relation-aware RelTracktor by plugging in the REM. To do so, we modify the regression head of the tracker to take the concatenated relation-appearance feature instead of just the appearance feature as the input. The rest of the tracker remains unchanged.

We use Xavier initialization \cite{xavier10init} for the relation encoding module. We also initialize the modified regression head from the backbone tracker. We then jointly train the modified regression head and the relational module. To do so, we randomly sample $T=10$ consecutive frames from MOT17/MOT20 datasets, compute relation embeddings and feed them into the regression head together with appearance features to refine bounding boxes at time step $T+1$. We train for 50 epochs using the Adam optimizer \cite{kingma2017adam} with a learning rate of $0.0001$ while setting $d_{th}=15$ to build relational graphs. We choose $F=128$ for a dimension of the relation embedding vectors. Generalized intersection over union \cite{Rezatofighi_2018_CVPR} is used as a loss function. We highlight that only the relational module and the regression head are trained, while the rest of the model is kept as is.

\subsection{Datasets and evaluation metrics}

The MOT17 benchmark consists of 7 train and 7 test sequences, which contain pedestrians with annotated full-body bounding boxes. It also specifies the degree of visibility for each annotated instance in the train split. The MOT20 benchmark contains 4 train and 4 test sequences of moving pedestrians in unconstrained environments with bounding boxes, covering the visible part of the objects.

Following \cite{tracktor_2019_ICCV}, we evaluate the multi-object tracking quality in a public detection setting. Such an evaluation protocol allows for a fair comparison with other methods. We employ standard the MOT-metrics \cite{bernardin2008clearmot} and the HOTA metric \cite{luiten2020IJCV} as an indicator of the overall performance. We additionally analyze HOTA over different localization thresholds. In a scene that contains a lot of dense interactions, a higher HOTA under low localization thresholds indicates more robust tracking of bodily interacting objects.

\subsection{Discovering relations}

We start by testing the ability of the relation encoding module to catch inter-object relations in the scene. As the MOT-benchmark does not provide explicit annotation for the relations, we conduct a qualitative study. In particular, from the relation-aware tracker we compute relation-importance weights (Equation \ref{eq:rel_imp}) for the objects in each time-step. We then visualize the top relations in Figure \ref{fig:discovered_rel} by the same color.

As can be seen, objects moving in a group tend to have stronger relations. We also observe that relations often are formed between coherently moving objects, even if they are not a part of what we would assess as a group (on our social experience). Such relations are inevitable as we do not employ social rules into the model, but still these assessed relations are useful as the trajectories of such ad-hoc groups can still be explained together.

\begin{table}
\vspace{1mm}
\center
\tabcolsep=0.11cm

    \resizebox{\columnwidth}{!}{
    \begin{tabular}{c l c c c c c c c c}
     \toprule
          & Method & HOTA $\uparrow$ & IDF1 $\uparrow$ & MOTA $\uparrow$ & MOTP $\uparrow$ & MT $\uparrow$ & ML $\downarrow$ \\ [0.5ex] 
     \midrule
     
     \parbox[t]{3mm}{\multirow{6}{*}{\rotatebox[origin=c]{90}{\footnotesize{MOT17}}}}
          & \textbf{RelTracktor} \textit{ (Ours)} & \textbf{45.8} & \textbf{56.5} & \textbf{57.2} & \textbf{79.0} & 21.9 & \textbf{34.3}\\
          & Tracktor \cite{tracktor_2019_ICCV} & 44.8 & 55.1 & 56.3 & 78.8 & 21.1 & 35.3 \\
          & deepMOT \cite{xu2020train} & 42.4 & 53.8 & 53.7 & 77.2 & 19.4 & 36.6 \\
          & eHAF \cite{sheng2019ehaf} & - & 54.7 & 51.8 & - & \textbf{23.4} & 37.9 \\
          & FWT \cite{henschel2017fwt} & - & 47.6 & 51.3 & - & 21.4 & 35.2 \\
          & jCC \cite{keuper2020jcc} & - & 54.5 & 51.2 & - & 20.9 & 37.0 \\
     \midrule
     
     \parbox[t]{3mm}{\multirow{6}{*}{\rotatebox[origin=c]{90}{\hspace{12mm} \footnotesize{MOT20}}}}
          & \textbf{RelTracktor} \textit{ (Ours)} & \textbf{43.4} & \textbf{53.0} & \textbf{54.1} & 79.2 & \textbf{36.7} & \textbf{22.6} \\
          & Tracktor \cite{tracktor_2019_ICCV} & 42.1 & 52.7 & 52.6 & \textbf{79.9} & 29.4 & 26.7 \\
          & SORT20 \cite{bewley2016sort} & 36.1 & 45.1 & 42.7 & 78.5 & 16.7 & 26.2 \\
          \bottomrule
    \end{tabular}}
\caption{Performance comparison on MOT17 and MOT20. The relation-aware RelTracktor model outperforms the baseline model with no relations on both benchmarks. RelTracktor improves the conventional tracker in various aspects as indicated by a range of metrics, while the HOTA improvement indicates the rise in the overall tracking quality.}
\label{tab:mot}
\end{table}

\subsection{Relation-aware tracking-by-regression}

We compare the relation-aware RelTracktor versus the baseline method from \cite{tracktor_2019_ICCV} and other trackers. We run the tracker on the test subset of MOT benchmarks and submit results to the evaluation server. Results are presented in Table \ref{tab:mot}.

On the MOT17-benchmark, the relation-aware tracker shows an improvement in all metrics compared to the Tracktor baseline. In particular, a higher IDF1 score indicates that our model robustly preserves the identities of the objects throughout the sequence, while also providing more accurate localization as indicated by the MOTP score. Specifically, we observed that when the baseline model fails for close objects in a group, the plain tracker drifts to follow the non-occluded member of the group and loses the target. The relation aware extension of the tracker, on the other hand, is more robust in such scenarios. We validate this by investigating failure cases and by analyzing HOTA values over different localization thresholds in supplementary materials.

On the MOT20-benchmark, the relation-aware tracker demonstrates 1.3\% increase in the overall HOTA score. Although the baseline tracker provides slightly higher localization precision as indicated by MOTP score, its relation-aware extension is much more robust and is able to track targets longer as indicated by the higher percentage of mostly tracked objects (MT).

\subsection{Tracking-by-relations}

\begin{figure*}[t]
  \centering
    \includegraphics[width=0.98\linewidth]{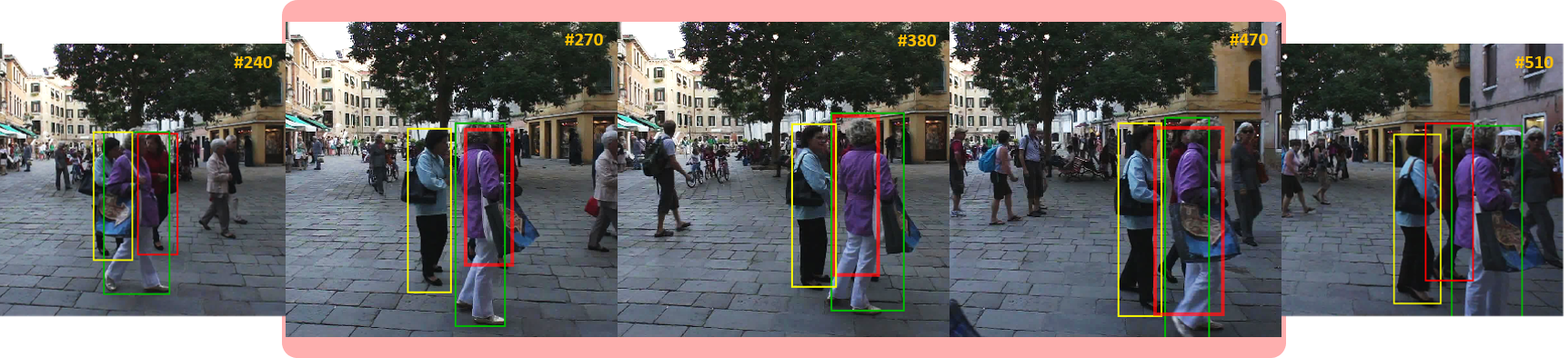}
  \caption{Tracking occluded object (red bounding box) by its relations with neighbors. The position of the object in the shaded region is recovered from its relation embedding.}
  \label{fig:tbr}
\end{figure*}

We next demonstrate that the position of out of sight objects can be recovered based purely on relation cues. To do so, we select an instance, which undergoes occlusion while moving in a group of related objects. We then apply the approach described in Section \ref{sec:model}. Practically, we do not kill the trajectories of non-visible objects, but continue to predict them from relational cues. 

Figure \ref{fig:tbr} demonstrate qualitative results. As can be seen, the position of the occluded object can be approximately recovered from the group relations. It indicates that relation embeddings can be learned to encode the geometric prior about the relative positions of the objects. We provide more qualitative examples in supplementary materials.
\section{Discussion}
\label{sec:discussion}

In this work, we demonstrate that learning inter-object relations is important for robust multi-object tracking. We develop a plug-in relation encoding module, which encodes relations by running a message passing over a spatial-temporal graph of tracked instances. 

We experimentally demonstrate that extending a backbone multi-object tracker with REM improves tracking quality. We also investigate the ability of the proposed method to track heavily occluded objects based on the relational cues, when appearance information is unreliable. Our experiments suggest that relational information is important and should not be left out from the analysis.

We suppose that REM would be the most useful in problems, where video analysis of crowded scenes with regular occlusions is required. For example, multi-object tracking in crowded streets, underground, airports, and railway stations. Moreover, one can use the proposed relation encoding module not only for tracking but also for action recognition or scene anomaly detection. We leave these questions for further research.

\bibliography{egbib}

\begin{thebibliography}{34}
\providecommand{\natexlab}[1]{#1}
\providecommand{\url}[1]{\texttt{#1}}
\expandafter\ifx\csname urlstyle\endcsname\relax
  \providecommand{\doi}[1]{doi: #1}\else
  \providecommand{\doi}{doi: \begingroup \urlstyle{rm}\Url}\fi

\bibitem[Bergmann et~al.(2019)Bergmann, Meinhardt, and
  Leal{-}Taix{\'{e}}]{tracktor_2019_ICCV}
Philipp Bergmann, Tim Meinhardt, and Laura Leal{-}Taix{\'{e}}.
\newblock Tracking without bells and whistles.
\newblock In \emph{The IEEE International Conference on Computer Vision
  (ICCV)}, October 2019.

\bibitem[Bernardin and Stiefelhagen(2008)]{bernardin2008clearmot}
Keni Bernardin and Rainer Stiefelhagen.
\newblock Evaluating multiple object tracking performance: The clear mot
  metrics.
\newblock \emph{J. Image Video Process.}, 2008, January 2008.
\newblock ISSN 1687-5176.
\newblock \doi{10.1155/2008/246309}.

\bibitem[Bewley et~al.(2016)Bewley, Ge, Ott, Ramos, and
  Upcroft]{bewley2016sort}
Alex Bewley, Zongyuan Ge, Lionel Ott, Fabio Ramos, and Ben Upcroft.
\newblock Simple online and realtime tracking.
\newblock In \emph{2016 IEEE International Conference on Image Processing
  (ICIP)}, pages 3464--3468, 2016.
\newblock \doi{10.1109/ICIP.2016.7533003}.

\bibitem[Cai et~al.(2019)Cai, Ge, Liu, Cai, Cham, Yuan, and
  Thalmann]{cai2019exploiting}
Yujun Cai, Liuhao Ge, Jun Liu, Jianfei Cai, Tat-Jen Cham, Junsong Yuan, and
  Nadia~Magnenat Thalmann.
\newblock Exploiting spatial-temporal relationships for 3d pose estimation via
  graph convolutional networks.
\newblock In \emph{Proceedings of the IEEE International Conference on Computer
  Vision}, pages 2272--2281, 2019.

\bibitem[Dendorfer et~al.(2020)Dendorfer, Rezatofighi, Milan, Shi, Cremers,
  Reid, Roth, Schindler, and Leal-Taixé]{dendorfer2020mot20}
Patrick Dendorfer, Hamid Rezatofighi, Anton Milan, Javen Shi, Daniel Cremers,
  Ian Reid, Stefan Roth, Konrad Schindler, and Laura Leal-Taixé.
\newblock Mot20: A benchmark for multi object tracking in crowded scenes, 2020.

\bibitem[Girshick et~al.(2018)Girshick, Radosavovic, Gkioxari, Doll\'{a}r, and
  He]{Detectron2018}
Ross Girshick, Ilija Radosavovic, Georgia Gkioxari, Piotr Doll\'{a}r, and
  Kaiming He.
\newblock Detectron.
\newblock \url{https://github.com/facebookresearch/detectron}, 2018.

\bibitem[Glorot and Bengio(2010)]{xavier10init}
Xavier Glorot and Yoshua Bengio.
\newblock Understanding the difficulty of training deep feed forward neural
  networks.
\newblock In Yee~Whye Teh and Mike Titterington, editors, \emph{Proceedings of
  the Thirteenth International Conference on Artificial Intelligence and
  Statistics}, volume~9 of \emph{Proceedings of Machine Learning Research},
  pages 249--256. PMLR, 2010.

\bibitem[Henschel et~al.(2017)Henschel, Leal-Taixé, Cremers, and
  Rosenhahn]{henschel2017fwt}
Roberto Henschel, Laura Leal-Taixé, Daniel Cremers, and Bodo Rosenhahn.
\newblock Improvements to frank-wolfe optimization for multi-detector
  multi-object tracking.
\newblock \emph{CVPR}, 05 2017.

\bibitem[Hu et~al.(2018)Hu, Gu, Zhang, Dai, and Wei]{Hu2018reldet}
Han Hu, Jiayuan Gu, Zheng Zhang, Jifeng Dai, and Yichen Wei.
\newblock Relation networks for object detection.
\newblock In \emph{CVPR}, pages 3588--3597, 2018.

\bibitem[Keuper et~al.(2016)Keuper, Tang, Zhongjie, Andres, Brox, and
  Schiele]{keuper2016multicut}
Margret Keuper, Siyu Tang, Yu~Zhongjie, Bjoern Andres, Thomas Brox, and Bernt
  Schiele.
\newblock A multi-cut formulation for joint segmentation and tracking of
  multiple objects, 2016.

\bibitem[Keuper et~al.(2020)Keuper, Tang, Andres, Brox, and
  Schiele]{keuper2020jcc}
Margret Keuper, Siyu Tang, Bjoern Andres, Thomas Brox, and Bernt Schiele.
\newblock Motion segmentation multiple object tracking by correlation
  co-clustering.
\newblock \emph{IEEE Transactions on Pattern Analysis and Machine
  Intelligence}, 42\penalty0 (1):\penalty0 140--153, 2020.

\bibitem[Kingma and Ba(2017)]{kingma2017adam}
Diederik~P. Kingma and Jimmy Ba.
\newblock Adam: A method for stochastic optimization, 2017.

\bibitem[Lee et~al.(2019)Lee, Lee, Kim, Kosiorek, Choi, and Teh]{lee2019set}
Juho Lee, Yoonho Lee, Jungtaek Kim, Adam Kosiorek, Seungjin Choi, and Yee~Whye
  Teh.
\newblock Set transformer: A framework for attention-based
  permutation-invariant neural networks.
\newblock In \emph{Proceedings of the 36th International Conference on Machine
  Learning}, pages 3744--3753, 2019.

\bibitem[Lu et~al.(2020)Lu, Rathod, Votel, and Huang]{Lu_2020_CVPR}
Zhichao Lu, Vivek Rathod, Ronny Votel, and Jonathan Huang.
\newblock Retinatrack: Online single stage joint detection and tracking.
\newblock In \emph{Proceedings of the IEEE/CVF Conference on Computer Vision
  and Pattern Recognition (CVPR)}, June 2020.

\bibitem[Luiten et~al.(2020)Luiten, Osep, Dendorfer, Torr, Geiger,
  Leal-Taix{\'e}, and Leibe]{luiten2020IJCV}
Jonathon Luiten, Aljosa Osep, Patrick Dendorfer, Philip Torr, Andreas Geiger,
  Laura Leal-Taix{\'e}, and Bastian Leibe.
\newblock Hota: A higher order metric for evaluating multi-object tracking.
\newblock \emph{International Journal of Computer Vision}, pages 1--31, 2020.

\bibitem[Materzynska et~al.(2020)Materzynska, Xiao, Herzig, Xu, Wang, and
  Darrell]{materzynska2020somethingelse}
Joanna Materzynska, Tete Xiao, Roei Herzig, Huijuan Xu, Xiaolong Wang, and
  Trevor Darrell.
\newblock Something-else: Compositional action recognition with
  spatial-temporal interaction networks.
\newblock In \emph{CVPR}, 2020.

\bibitem[Milan et~al.(2016)Milan, Leal-Taix{\'e}, Reid, Roth, and
  Schindler]{Milan2016MOT16AB}
Anton Milan, Laura Leal-Taix{\'e}, Ian~D. Reid, Stefan Roth, and Konrad
  Schindler.
\newblock Mot16: A benchmark for multi-object tracking.
\newblock \emph{ArXiv}, abs/1603.00831, 2016.

\bibitem[Narasimhan et~al.(2018)Narasimhan, Lazebnik, and
  Schwing]{Narasimhan2018OutOT}
Medhini Narasimhan, S.~Lazebnik, and A.~Schwing.
\newblock Out of the box: Reasoning with graph convolution nets for factual
  visual question answering.
\newblock In \emph{NeurIPS}, 2018.

\bibitem[Ren et~al.(2017)Ren, He, Girshick, and Sun]{ren2017faster}
Shaoqing Ren, Kaiming He, Ross Girshick, and Jian Sun.
\newblock Faster {R-CNN}: Towards real-time object detection with region
  proposal networks.
\newblock \emph{IEEE Transactions on Pattern Analysis and Machine
  Intelligence}, 39\penalty0 (6):\penalty0 1137--1149, 2017.

\bibitem[Rezatofighi et~al.(2019)Rezatofighi, Tsoi, Gwak, Sadeghian, Reid, and
  Savarese]{Rezatofighi_2018_CVPR}
Hamid Rezatofighi, Nathan Tsoi, JunYoung Gwak, Amir Sadeghian, Ian Reid, and
  Silvio Savarese.
\newblock Generalized intersection over union.
\newblock 2019.

\bibitem[Ristani et~al.(2016)Ristani, Solera, Zou, Cucchiara, and
  Tomasi]{ristani2016idf}
Ergys Ristani, Francesco Solera, Roger~S. Zou, R.~Cucchiara, and Carlo Tomasi.
\newblock Performance measures and a data set for multi-target, multi-camera
  tracking.
\newblock \emph{European Conference on Computer Vision}, pages 17--35, 2016.

\bibitem[Sheng et~al.(2019)Sheng, Zhang, Chen, Xiong, and Zhang]{sheng2019ehaf}
Hao Sheng, Yang Zhang, Jiahui Chen, Zhang Xiong, and Jun Zhang.
\newblock Heterogeneous association graph fusion for target association in
  multiple object tracking.
\newblock \emph{IEEE Transactions on Circuits and Systems for Video
  Technology}, 29\penalty0 (11):\penalty0 3269--3280, 2019.

\bibitem[{Tang} et~al.(2015){Tang}, {Andres}, {Andriluka}, and
  {Schiele}]{tang2015subgraph}
S.~{Tang}, B.~{Andres}, M.~{Andriluka}, and B.~{Schiele}.
\newblock Subgraph decomposition for multi-target tracking.
\newblock In \emph{2015 IEEE Conference on Computer Vision and Pattern
  Recognition (CVPR)}, pages 5033--5041, 2015.

\bibitem[Tang et~al.(2016)Tang, Andres, Andriluka, and Schiele]{tang2016multi}
Siyu Tang, Bjoern Andres, Mykhaylo Andriluka, and Bernt Schiele.
\newblock Multi-person tracking by multicut and deep matching.
\newblock In \emph{Computer Vision -- ECCV 2016 Workshops}, pages 100--111,
  2016.

\bibitem[Tang et~al.(2017)Tang, Andriluka, Andres, and
  Schiele]{tang2017multiple}
Siyu Tang, Mykhaylo Andriluka, Bjoern Andres, and Bernt Schiele.
\newblock Multiple people tracking by lifted multicut and person
  re-identification.
\newblock In \emph{CVPR}, pages 3701--3710, 2017.

\bibitem[Vaswani et~al.(2017)Vaswani, Shazeer, Parmar, Uszkoreit, Jones, Gomez,
  Kaiser, and Polosukhin]{Vaswani2017attention}
Ashish Vaswani, Noam Shazeer, Niki Parmar, Jakob Uszkoreit, Llion Jones,
  Aidan~N Gomez, \L~ukasz Kaiser, and Illia Polosukhin.
\newblock Attention is all you need.
\newblock In \emph{Advances in Neural Information Processing Systems},
  volume~30, 2017.

\bibitem[Veli{\v{c}}kovi{\'{c}} et~al.(2018)Veli{\v{c}}kovi{\'{c}}, Cucurull,
  Casanova, Romero, Li{\`{o}}, and Bengio]{velickovic2018graph}
Petar Veli{\v{c}}kovi{\'{c}}, Guillem Cucurull, Arantxa Casanova, Adriana
  Romero, Pietro Li{\`{o}}, and Yoshua Bengio.
\newblock {Graph Attention Networks}.
\newblock \emph{International Conference on Learning Representations}, 2018.

\bibitem[Voigtlaender et~al.(2019)Voigtlaender, Krause, Osep, Luiten, Sekar,
  Geiger, and Leibe]{Voigtlaender19CVPR_MOTS}
Paul Voigtlaender, Michael Krause, Aljosa Osep, Jonathon Luiten, Berin
  Balachandar~Gnana Sekar, Andreas Geiger, and Bastian Leibe.
\newblock {MOTS}: Multi-object tracking and segmentation.
\newblock In \emph{CVPR}, 2019.

\bibitem[Wang et~al.(2019)Wang, Zheng, Liu, and Wang]{wang2019towards}
Zhongdao Wang, Liang Zheng, Yixuan Liu, and Shengjin Wang.
\newblock Towards real-time multi-object tracking.
\newblock \emph{arXiv preprint arXiv:1909.12605}, 2019.

\bibitem[Wu et~al.(2021)Wu, Pan, Chen, Long, Zhang, and Yu]{wu2021gnnsurvey}
Zonghan Wu, Shirui Pan, Fengwen Chen, Guodong Long, Chengqi Zhang, and
  Philip~S. Yu.
\newblock A comprehensive survey on graph neural networks.
\newblock \emph{IEEE Transactions on Neural Networks and Learning Systems},
  32\penalty0 (1):\penalty0 4--24, 2021.

\bibitem[Xu et~al.(2020)Xu, Osep, Ban, Horaud, Leal-Taix{\'e}, and
  Alameda-Pineda]{xu2020train}
Yihong Xu, Aljosa Osep, Yutong Ban, Radu Horaud, Laura Leal-Taix{\'e}, and
  Xavier Alameda-Pineda.
\newblock How to train your deep multi-object tracker.
\newblock In \emph{Proceedings of the IEEE/CVF Conference on Computer Vision
  and Pattern Recognition}, pages 6787--6796, 2020.

\bibitem[Zhang et~al.(2020)Zhang, Wang, Wang, Zeng, and Liu]{zhang2020fair}
Yifu Zhang, Chunyu Wang, Xinggang Wang, Wenjun Zeng, and Wenyu Liu.
\newblock Fairmot: On the fairness of detection and re-identification in
  multiple object tracking.
\newblock \emph{arXiv preprint arXiv:2004.01888}, 2020.

\bibitem[Zhou et~al.(2019)Zhou, Wang, and Kr{\"a}henb{\"u}hl]{zhou2019objects}
Xingyi Zhou, Dequan Wang, and Philipp Kr{\"a}henb{\"u}hl.
\newblock Objects as points.
\newblock In \emph{arXiv preprint arXiv:1904.07850}, 2019.

\bibitem[Zhou et~al.(2020)Zhou, Koltun, and
  Kr{\"a}henb{\"u}hl]{zhou2020tracking}
Xingyi Zhou, Vladlen Koltun, and Philipp Kr{\"a}henb{\"u}hl.
\newblock Tracking objects as points.
\newblock \emph{ECCV}, 2020.

\end{thebibliography}

\appendix
\clearpage

\begin{figure*}[h!]
\vspace{1mm}
  \centering
    \includegraphics[width=0.98\linewidth]{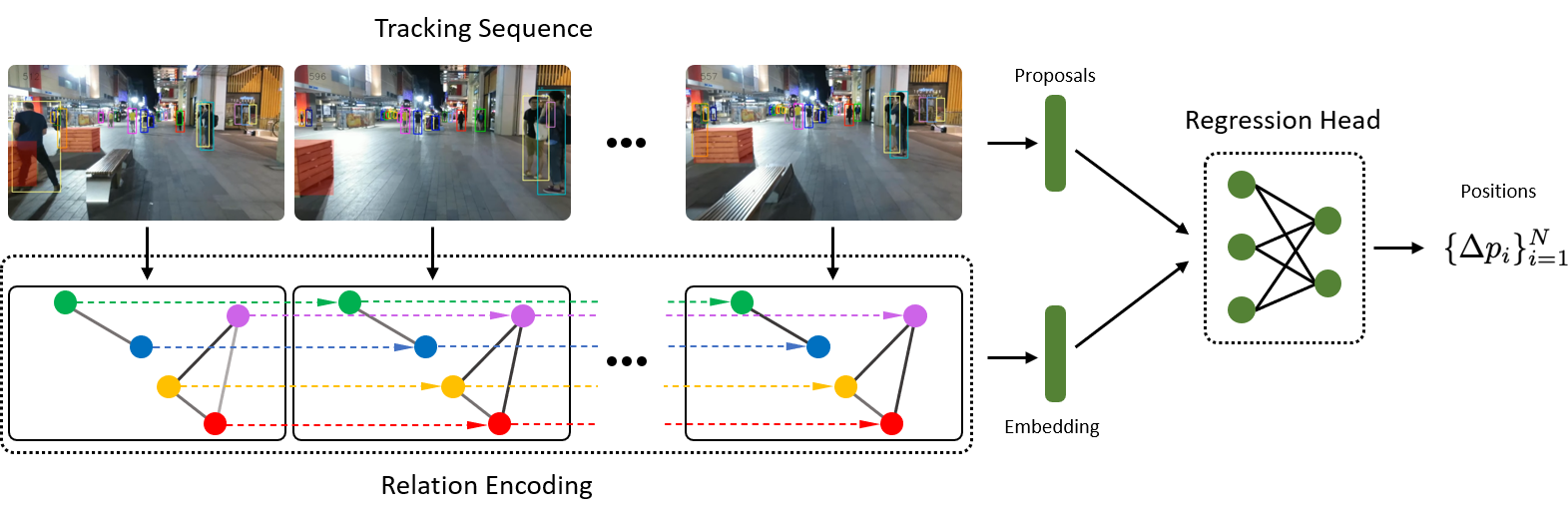}
  \caption{The relational graph for the REM is built on top of the tracked objects. The constructed graph is used to compute relation embeddings, which guide the regression head of a backbone tracker.}
  \label{fig:overall}
\end{figure*}

\section{Utilizing relations for tracking}

\paragraph{Relation-aware tracking-by-regression} We illustrate a pipeline of a tracking-by-regression extended with relation reasoning in Figure \ref{fig:overall}. The graph is built dynamically from the tracked instances. When the object enters or leaves the scene, the corresponding node in the relational graph is updated. 

\paragraph{Tracking-by-relations} As the relation embeddings contain the prior about object's positions, the location of the object can be approximately recovered from the relational cues. To that end, we pass the relation embeddings through an output MLP $:\mathbbm{R}^{F} \rightarrow \mathbbm{R}^{4}$, which predicts the coordinates of the object. 

We set the dimension of the relation embedding $r_i^{t} \in \mathbbm{R}^{128}$. Then, the output MLP is a two layer feed forward network $\texttt{128} \rightarrow \texttt{64} \rightarrow \texttt{4}$ with LeakyReLU non-linearities. The output MLP is jointly trained with the main model by minimizing the generalized intersection over union \cite{Rezatofighi_2018_CVPR} with respect to the ground truth bounding boxes.

\section{Experiments}

\paragraph{Ablations studies}

We conduct an ablation study on the MOT17-FRCNN train split to investigate an impact of the distance used to build the relational graph $d_{th}$, with higher $d_{th}$ resulting in a relational graph of a larger spatial extent. Since the relation-aware model is fine-tuned for the tracking task, we also compare against the fine-tuned baseline method from \cite{tracktor_2019_ICCV}.

As can be seen in Table \ref{tab:ablations}, the relation-aware tracker outperforms the baseline method and its fine-tuned version. We observed that using the relational graphs of a larger spatial extent generally results in higher performance. Intuitively, lower $d_{th}$ restricts to encode relation only in small groups, missing the possible distant connections. At the same time, a very large $d_{th}$ leads to a slight decrease in performance. We attribute it to the fact that in very big relational graphs, formed relations are less sharp because of the attention aggregation mechanism.

\paragraph{Analyzing HOTA}

\begin{figure*}[t!]
\vspace{1mm}
  \centering
    \includegraphics[width=0.98\linewidth]{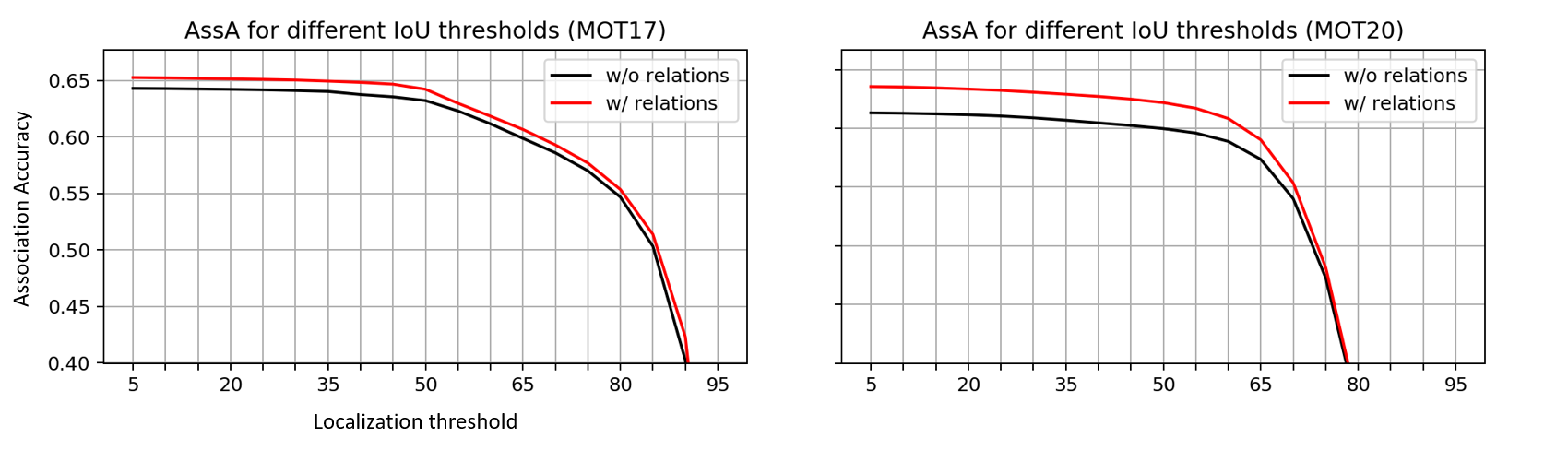}
  \caption{Association Accuracy (AssA) of the relation-aware and the baseline tracker over different localization thresholds.}
  \label{fig:assa}
\end{figure*}

\begin{table}[t]
\centering
\begin{tabular}{@{}cccc@{}}
    \toprule
    Method & IDF1 & MOTA & MOTP \\ \midrule
    \textit{Base} & 65.0 & 61.7 & 89.4 \\
    \textit{Base-tuned} & 66.7 & 61.8 & 89.5 \\
    \textit{Rel} $d_{th}=\textit{5}$ & 66.8 & 61.9 & 89.4 \\
    \textit{Rel} $d_{th}=\textit{10}$ & 66.9 & 61.9 & 89.5 \\
    \textit{Rel} $d_{th}=\textit{20}$ & \textbf{67.1} & \textbf{62.0} & \textbf{89.5} \\
    \textit{Rel} $d_{th}=\textit{30}$ & \textbf{67.1} & \textbf{62.0} & \textbf{89.5} \\ 
    \textit{Rel} $d_{th}=\textit{40}$ & 66.9 & 61.9 & 89.4 \\ 
    \bottomrule
    \end{tabular}
    \vspace{2mm}
    \caption{Comparing the baseline and relation-aware models over different distance threshold used to build relational graph. \textit{Rel} $d_{th}=\textit{20}$ stands for the relation-aware tracker with distance threshold of 20 to build relational graph.}
\label{tab:ablations}
\end{table}

To investigate the ability of the relation-aware model to provide robust tracking in the dense scenes, we decompose and analyze HOTA metrics over various localization thresholds $\alpha$. From \cite{luiten2020IJCV}:

\begin{equation}
    HOTA_{\alpha} = \sqrt{DetA_{\alpha} \cdot AssA_{\alpha}}
\end{equation}

where $DetA_{\alpha}, AssA_{\alpha}$ stand for detection and association accuracy for a given $\alpha$. Since the classification part responsible for the quality of detection remains unchanged, we only analyze the association accuracy. Association accuracy for tracking-by-regression methods measures the ability of the regression head of the tracker to preserve identities frame by frame. We plot $AssA_{\alpha}$ for the baseline and relation-aware trackers at Figure \ref{fig:assa}.

Low localization thresholds permit the association of loose bounding boxes. It can deteriorate the association quality when predicted bounding boxes are densely overlapped, which is a common case in crowded scenarios. Thus, the higher association accuracy of relation-aware tracker (Figure \ref{fig:assa}) under low localization thresholds indicates the better ability to preserve identities of densely interacting objects.

\begin{figure*}[t!]
\vspace{1mm}
  \centering
    \includegraphics[width=0.98\linewidth]{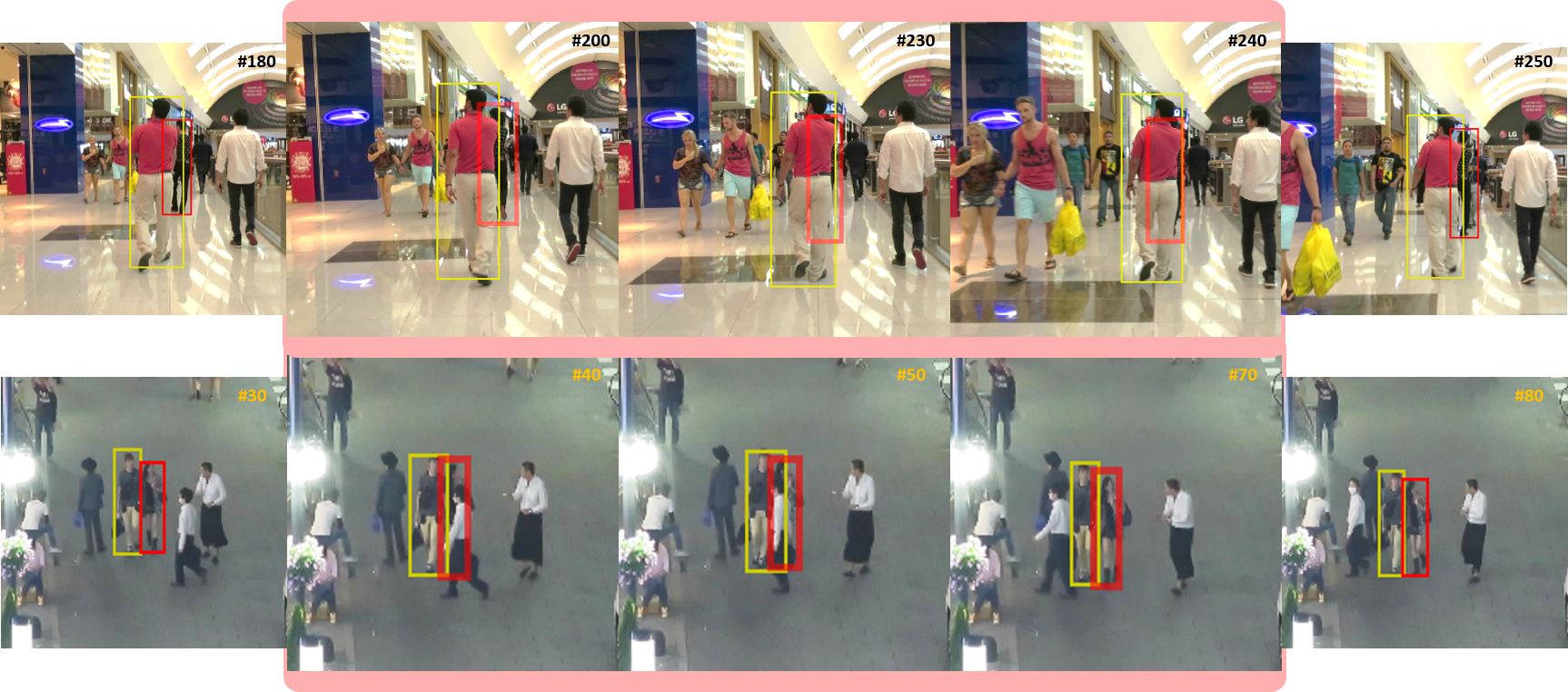}
  \caption{Tracking occluded object (red bounding box) by its relations with neighbors. The position of the object in the shaded region is recovered from its relation embedding.}
  \label{fig:tbr_supp}
\end{figure*}

\paragraph{Qualitative examples of tracking-by-relations} We provide more qualitative examples of the tracking-by-relation approach in Figure \ref{fig:tbr_supp}.

\paragraph{Failure cases} When inspecting the per-sequence results of the relation-aware tracker, we observed the improvement of tracking quality on a range of sequences compared to the baseline. The modest performance gain was on the sequences with severe camera motion. We attribute it to the fact that the global camera motion enforces the abrupt shifts in object trajectories, which hinders the formation of meaningful relations.

\end{document}